\documentclass[10pt,twocolumn,letterpaper]{article}

\usepackage{iccv}              

\usepackage[utf8]{inputenc} 
\usepackage[T1]{fontenc}    
\usepackage{url}            
\usepackage{booktabs}       
\usepackage{amsfonts}       
\usepackage{nicefrac}       
\usepackage{colortbl} 
\usepackage{microtype}      
\usepackage{xcolor}         

\usepackage{xcolor,soul,framed} 

\usepackage{array}
\usepackage{mdwmath}
\usepackage{mdwtab}
\usepackage{eqparbox}
\usepackage{url}
\usepackage{multirow}
\usepackage{graphicx}
\usepackage{amsfonts}
\usepackage{amsmath}
\usepackage{amsthm}
\usepackage{amssymb}
\usepackage{graphicx}
\usepackage{multirow}

\definecolor{lightblue}{rgb}{0.9, 0.95, 1.0}
\definecolor{mygreen}{rgb}{0.01, 0.5, 0.01}
\definecolor{myred}{rgb}{0.8, 0.01, 0.01}

\newtheorem{theorem}{Theorem}

\definecolor{iccvblue}{rgb}{0.21,0.49,0.74}
\usepackage[pagebackref,breaklinks,colorlinks,allcolors=iccvblue]{hyperref}
\usepackage{graphicx}
\usepackage{xcolor}

\def\paperID{\#1} 
\def\confName{ICCV}
\def\confYear{2025}

\title{AdeptHEQ-FL: Adaptive Homomorphic Encryption for Federated Learning of Hybrid Classical-Quantum Models with Dynamic Layer Sparing}

\author{
    \textbf{Md Abrar Jahin}$^{1*}$,
    \textbf{Taufikur Rahman Fuad}$^{2}$,
    \textbf{M. F. Mridha}$^{3*}$,
    \textbf{Nafiz Fahad}$^{4}$,
    \textbf{Md. Jakir Hossen}$^{4*}$\\
        $^{1}$University of Southern California\\
        $^{2}$Islamic University of Technology\\
        $^{3}$American International University-Bangladesh\\
        $^{4}$Multimedia University\\
        \texttt{jahin@usc.edu},\quad \texttt{taufikur@iut-dhaka.edu},\\ \texttt{firoz.mridha@aiub.edu},\quad \texttt{jakir.hossen@mmu.edu.my}
}

\begin{document}
\maketitle

\begin{abstract}
Federated Learning (FL) faces inherent challenges in balancing model performance, privacy preservation, and communication efficiency, especially in non-IID decentralized environments. Recent approaches either sacrifice formal privacy guarantees, incur high overheads, or overlook quantum-enhanced expressivity. We introduce AdeptHEQ-FL, a unified hybrid classical-quantum FL framework that integrates (i) a hybrid CNN-PQC architecture for expressive decentralized learning, (ii) an adaptive accuracy-weighted aggregation scheme leveraging differentially private validation accuracies, (iii) selective homomorphic encryption (HE) for secure aggregation of sensitive model layers, and (iv) dynamic layer-wise adaptive freezing to minimize communication overhead while preserving quantum adaptability. We establish formal privacy guarantees, provide convergence analysis, and conduct extensive experiments on the CIFAR-10, SVHN, and Fashion-MNIST datasets. AdeptHEQ-FL achieves a $\approx 25.43\%$ and $\approx 14.17\%$ accuracy improvement over Standard-FedQNN and FHE-FedQNN, respectively, on the CIFAR-10 dataset. Additionally, it reduces communication overhead by freezing less important layers, demonstrating the efficiency and practicality of our privacy-preserving, resource-aware design for FL. Our code is publicly available at: \url{https://github.com/Abrar2652/QML-FL}.
\end{abstract}

\section{Introduction}
\label{sec:introduction}
Federated Learning (FL) has emerged as a transformative paradigm for collaborative Machine Learning (ML), allowing decentralized devices to train a shared model without centralizing sensitive data \cite{mcmahan2017communication, kairouz2021advances}. This approach is crucial for privacy-sensitive applications, such as personalized medicine, secure finance, and the Internet of Things (IoT), where data privacy and resource constraints are critical. However, effectively deploying FL is challenged by a triad of issues: statistical heterogeneity from non-Identical and non-Independently Distributed (non-IID) data, privacy vulnerabilities despite data localization, and high communication and computational overheads \cite{mcmahan2017communication, kairouz2021advances}. Non-IID data among clients often hinders model performance and slows convergence. While FL inherently maintains some privacy by keeping data local, model updates remain vulnerable to attacks that can infer sensitive information. Frequent model exchanges between clients and servers amplify communication costs, especially in bandwidth-constrained environments. These interconnected issues require a unified solution to improve the robustness and scalability of FL.

Existing approaches often address these challenges in isolation, resulting in fragmented solutions. Quantum FL (QFL) utilizes quantum circuits to improve model expressivity 
\cite{innan_fedqnn_2024, liu_practical_2025}, but many frameworks overlook formal privacy guarantees or non-IID robustness 
\cite{innan_fedqnn_2024}. Privacy-preserving techniques, such as Differential Privacy (DP) 
\cite{ullah_quantum_2024} and Homomorphic Encryption (HE) 
\cite{yan_towards_2024}, protect data but compromise utility in non-IID settings or incur significant overhead 
\cite{dutta_federated_2024}. Efficiency-focused methods, like model compression 
\cite{guo_fedsign_2023}, reduce communication but seldom integrate quantum capabilities or comprehensively address privacy. This gap highlights the need for a unified framework that optimizes performance, privacy, and efficiency in a hybrid classical-quantum context.

To bridge these identified gaps, we propose \textbf{AdeptHEQ-FL}, a novel framework designed as a \textit{unified} solution. Where existing QFL approaches often lack formal privacy or non-IID robustness, AdeptHEQ-FL synergistically combines its hybrid classical-quantum architecture with adaptive accuracy-weighted aggregation (utilizing differentially private validation accuracies) to explicitly improve performance on non-IID data while improving model expressivity. To counter the significant overhead or utility degradation associated with many privacy-preserving techniques, especially in non-IID settings, AdeptHEQ-FL strategically employs HE (CKKS scheme) on critical final classical layers during aggregation, balancing strong privacy with computational feasibility, and further bolsters utility through its adaptive aggregation that prioritizes more accurate client models. Unlike efficiency-focused methods that typically neglect quantum capabilities or comprehensive privacy, our dynamic layer sparing mechanism is integrated to reduce communication and computation, specifically exempting quantum layers to preserve their crucial adaptability and ensuring the overall privacy-preserving nature of the framework is maintained. By holistically integrating these components, AdeptHEQ-FL provides a more comprehensive approach than existing fragmented solutions, aiming to concurrently optimize performance, privacy, and efficiency within a hybrid classical-quantum FL paradigm.

Our \textbf{primary contributions} are: 
\textbf{(i)} We introduce a novel \textit{adaptive aggregation mechanism} for FL that employs differentially private client validation accuracies and HE to effectively address non-IID data and ensure privacy.
\textbf{(ii)} We propose a \textit{hybrid classical-quantum architecture} integrating CNNs for feature extraction with PQCs to improve model expressivity in federated settings.
\textbf{(iii)} We develop an \textit{efficient dynamic layer sparing technique} that reduces communication overhead by adaptively freezing less impactful classical layers while preserving the adaptability of quantum layers.
\textbf{(iv)} We provide a theoretical convergence analysis for the proposed framework, accounting for adaptive aggregation, layer sparing, and privacy mechanisms.

\section{Related Works}
\label{sec:related_works}
FL enables collaborative model training across decentralized devices while prioritizing data privacy, yet faces challenges from non-IID, privacy vulnerabilities, and high communication costs \cite{mcmahan_communication-efficient_2017,kairouz2021advances}. Recent efforts explore quantum computing, privacy-preserving mechanisms, and efficiency optimizations, often addressing these issues in isolation. We critically review these efforts across four dimensions—quantum-improved FL, privacy preservation, communication efficiency, and adaptive/specialized approaches—identifying gaps that our AdeptHEQ-FL framework addresses through adaptive accuracy-weighted aggregation, classical-quantum hybridization, and formal convergence guarantees.

\subsection{Quantum FL}
Quantum FL (QFL) leverages quantum circuits to improve model expressivity. FedQNN \cite{innan_fedqnn_2024} employs QNNs and discusses secure data handling, but lacks formal privacy mechanisms like DP, leaving potential vulnerabilities unaddressed. Similarly, \cite{rofougaran_federated_2024,ullah_quantum_2024} integrate DP into QFL but overlook communication costs and provide no convergence proofs, limiting their robustness. FHE-FedQNN \cite{dutta_federated_2024} combines fully HE (FHE) with quantum circuits, reporting results on datasets like CIFAR-10 \cite{krizhevsky2009learning}, Brain MRI \cite{msoud_nickparvar_2021_brain_mri}, and PCOS \cite{Handa_2024_pcos}. Its uniform aggregation struggles with non-IID data, and FHE’s complexity leads to high communication overhead, a general concern in FHE-based FL approaches, rendering it impractical for edge devices. Its extension, MQFL-FHE, while leveraging hybrid quantum-FHE operations for multimodal tasks, still faces computational and communication inefficiencies. Theoretical studies like \cite{chehimi_foundations_2024} explore Quantum Neural Networks (QNN) for FL without empirical validation, while \cite{liu_practical_2025} demonstrates QFL experimentally but omits formal convergence analysis. These works highlight QFL’s potential but fail to unify privacy, efficiency, and theoretical rigor, gaps AdeptHEQ-FL addresses.

\subsection{Privacy-Preserving Techniques}
Privacy in FL often relies on DP or HE. DP-based methods \cite{chen_federated_2025,ullah_quantum_2024} add noise to updates, degrading accuracy in non-IID settings \cite{kairouz2021advances}. HE-based aggregation \cite{sebert_combining_2023,yan_towards_2024} ensures security but introduces significant computational overhead, limiting scalability. Hybrid approaches like ADPHE-FL \cite{wu_adphe-fl_2025} and others \cite{aziz_privacy_2024,zhang_secure_2024,aziz_exploring_2023} adaptively combine DP and HE to balance privacy and utility in classical FL, yet neglect quantum improvements and communication efficiency for non-IID data. Comparative analyses \cite{catalfamo_privacy-preserving_2025} evaluate DP versus HE but offer no solutions for non-IID challenges, underscoring the need for AdeptHEQ-FL’s quantum-aware, adaptive privacy framework.

\subsection{Efficiency in FL}
Efficiency-focused FL methods aim to reduce communication and computational costs. FedSIGN \cite{guo_fedsign_2023} employs sign-based compression to lower bandwidth and provides convergence analysis, but lacks quantum compatibility, restricting its applicability to classical settings. Multi-party computation (MPC) approaches \cite{chen_secure_2024,tran_novel_2023,kaminaga_mpcfl_2023} reduce communication overhead, yet often compromise accuracy in non-IID settings and ignore quantum integration, as seen in their classical focus. These methods highlight a trade-off between efficiency and performance that AdeptHEQ-FL mitigates through adaptive layer freezing and quantum-improved aggregation.

\subsection{Adaptive and Specialized Approaches}
Adaptive FL frameworks like \cite{sorbera_adaptive_2025} explore functional encryption for security but do not address learning dynamics or non-IID convergence. Quantum-safe FL \cite{yan_towards_2024} applies HE without tackling non-IID data or providing guarantees. Quantum-inspired methods \cite{bhatia_application_2025,tanbhir_quantum-inspired_2025} optimize computation via tensor networks or Quantum Key Distribution (QKD) but lack formal convergence guarantees for non-IID settings. Application-specific FL, such as for mental healthcare \cite{gupta_privacy-preserving_2024}, emphasizes both privacy and scalability, noting trade-offs such as longer training times for improved privacy. AdeptHEQ-FL distinguishes itself by integrating adaptive aggregation, quantum advantages, and rigorous convergence analysis for non-IID settings, offering a comprehensive solution.

\section{Methodology}
\label{sec:methodology}
\begin{figure*}[!ht]
    \centering
    \includegraphics[width=0.9\linewidth]{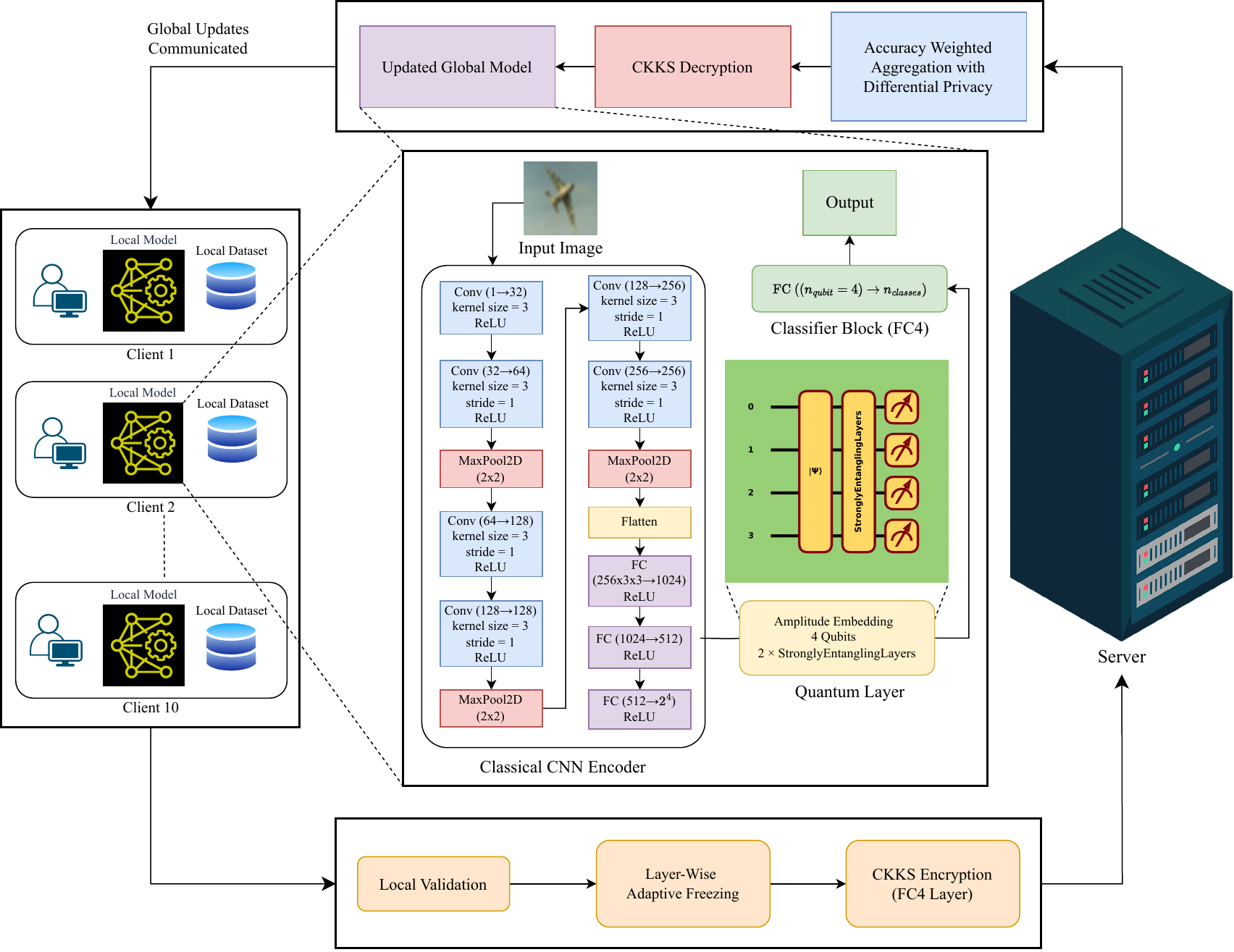}
    \caption{This flowchart provides an overview of the AdeptHEQ-FL framework and illustrates the multi-stage process, detailing the activities conducted on both the client and server sides. Each client independently executes a local training phase using a local dataset on the classical--quantum neural network. This is followed by local validation, adaptive layer freezing, and encryption of the final classifier layer. The encrypted local models are then sent to a central server for global aggregation, resulting in an improved federated model. After aggregation, the updated global model is communicated back to each client, where it is used as the local model for the next round. The \textcolor{myred}{dotted line} here indicates a more detailed version of the blocks in the diagram.}
    \label{fig:diagram}
\end{figure*}

AdeptHEQ-FL is a novel FL framework that integrates classical and quantum neural networks to address challenges such as non-IID, privacy preservation, and communication efficiency. By combining performance-based adaptive aggregation, layer-wise adaptive freezing, and DP, AdeptHEQ-FL improves model performance, reduces communication overhead, and ensures client privacy while maintaining compatibility with HE. A high-level overview of the complete system of AdeptHEQ-FL is presented in Figure~\ref{fig:diagram}.

\subsection{Problem Formulation}
\label{subsec:problem_formulation}

Consider an FL scenario with \( N \) clients, each possessing a local dataset \( \mathcal{D}_i \sim p_i(x,y) \) that may exhibit non-IID distributions. The global model parameters \( \theta = [\theta^c, \theta^q] \) consist of classical (\( \theta^c \)) and quantum (\( \theta^q \)) components. We define the learning objective as:
\begin{equation}
\min_{\theta} \sum_{i=1}^N \underbrace{\frac{|\mathcal{D}_i|}{\sum_{j=1}^N |\mathcal{D}_j|}}_{\text{fixed } w_i} \mathcal{L}_i(\theta; \mathcal{D}_i) 
\label{eq:objective}
\end{equation}
where the fixed aggregation weights  \( w_i \) proportionally reflect each client's dataset size, ensuring that clients with larger datasets contribute more significantly to the global model. While this formulation establishes a stable baseline objective using static weights, the actual aggregation process (detailed in Section~\ref{subsec:adaptive_aggregation}) employs dynamic weights  \( w_i^{(t)} \) derived from privatized validation accuracies to address non-IID challenges. The hybrid architecture simultaneously optimizes both quantum and classical parameters, maintaining regularization stability in classical components while allowing quantum layers to adapt freely to complex data patterns.

\subsection{Model Architecture}
The proposed AdeptHEQ-FL model integrates both classical and quantum computing techniques, leveraging the strengths of each. The following subsections discuss each of these components.

\subsubsection{Classical Component}
The classical component of the architecture is implemented as a CNN, which is a type of DL model that is particularly effective for analyzing grid-like data such as images. CNNs function by applying multiple layers of convolutional filters that extract localized features from the input image, including edges, textures, and shapes.

In this implementation, the CNN is composed of three sequential convolutional blocks. Each block consists of multiple convolutional layers, followed by a Rectified Linear Unit (ReLU) activation function and a max-pooling layer. The convolutional layers perform a mathematical operation known as a discrete convolution:
\begin{equation}
s(i,j) = (I \times K)(i,j) = \sum_m \sum_n I(i+m, j+n) K(m, n)
\end{equation}
where $I(i,j)$ represents the input image and $K(m,n)$ is a learnable filter (also called a kernel), this operation slides the kernel across the input image, producing a feature map that highlights the presence of specific patterns detected by the filter. The ReLU activation function is then applied element-wise to the feature maps, transforming the values according to: $\text{ReLU}(x) = \max(0, x)$. This introduces non-linearity into the model, enabling it to learn complex representations of the data. Following the activations, a max-pooling operation is applied to downsample the feature maps, reducing their spatial dimensions and controlling overfitting by summarizing the most prominent features. After the final convolutional block, the feature maps are flattened into a one-dimensional vector and passed through fully connected (dense) layers to produce a final classical feature representation: $f_{\text{CNN}}(x; \theta^c) \in \mathbb{R}^{2^n}$, where $n$ is the number of qubits in the quantum circuit (here, $2^4 = 16$).

\subsubsection{Quantum Component}
The core innovation in AdeptHEQ-FL is the incorporation of a PQC, which serves as a QNN for feature processing. Unlike classical networks that manipulate continuous or discrete numerical values, quantum circuits process data encoded into quantum states. In this work, the PQC operates on 4 qubits and consists of 2 layers of Strongly Entangling Layers — a widely-used ansatz in variational quantum algorithms (Figure \ref{fig:qml_circuit}) \cite{jahin2023qamplifynet,jahin_lorentz_2025}. The quantum circuit performs the following steps:

\begin{figure*}[!ht]
    \centering
    \includegraphics[width=1\linewidth]{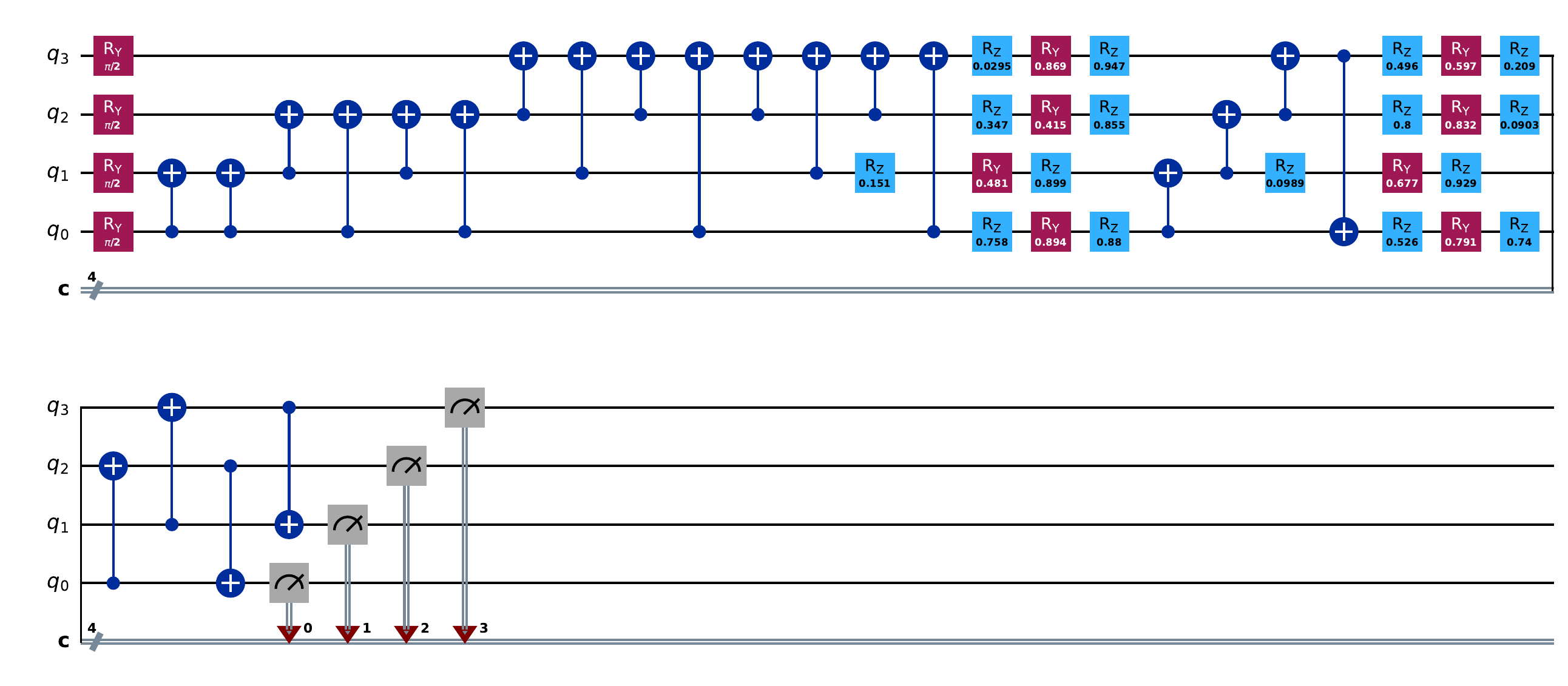}
    \caption{4-qubit 2-layered PQC of AdeptHEQ-FL comprising amplitude embedding, two Strongly Entangling Layers (parameterized $R_z$, $R_y$, $R_z$ rotations), CNOT-based entanglement, and projective measurements. The CNOT connectivity ensures full inter-qubit interaction within each layer.}
    \label{fig:qml_circuit}
\end{figure*}

\paragraph{Amplitude Embedding}\label{append:AE}
The output from the CNN, denoted by $f_{\text{CNN}}(x; \theta^c)$, is first encoded into the quantum circuit through amplitude embedding\footnote{\tiny\url{https://docs.pennylane.ai/en/stable/code/api/pennylane.AmplitudeEmbedding.html}}. This encoding maps a normalized classical vector $x \in \mathbb{R}^{2^n}$ into the amplitudes of a quantum state:
\begin{equation}
| \psi_x \rangle = \sum_{i=0}^{2^n-1} x_i |i\rangle
\end{equation}
where $|i\rangle$ represents the computational basis states of the qubit system. Amplitude embedding ensures that the sum of the squared amplitudes equals 1, maintaining a valid quantum state:
\begin{equation}
\sum_{i=0}^{2^n-1} |x_i|^2 = 1
\end{equation}

\paragraph{Strongly Entangling Layers}\label{append:SEL}
After the amplitude embedding step, the encoded quantum state undergoes a sequence of parameterized transformations and entangling operations, collectively termed as Strongly Entangling Layers. These layers are crucial for introducing both individual qubit rotations and inter-qubit correlations, allowing the quantum circuit to model complex feature interactions.

Each Strongly Entangling Layer\footnote{\tiny\url{https://docs.pennylane.ai/en/stable/code/api/pennylane.StronglyEntanglingLayers.html}} comprises two primary components. The first component involves a series of parameterized single-qubit rotation gates applied independently to each qubit. Specifically, for each qubit, a cascade of three rotations is performed in the order: $R_z(\theta^1) \rightarrow R_y(\theta^2) \rightarrow R_z(\theta^3)$. \( R_z(\theta_{k,i}^1) \) rotates the qubit about the Z-axis by an angle \( \theta_{k,i}^1 \), \( R_y(\theta_{k,i}^2) \) rotates about the Y-axis by \( \theta_{k,i}^2 \), followed again by \( R_z(\theta_{k,i}^3) \). The rotation operations are defined as:
\begin{equation}
\resizebox{0.9\hsize}{!}{$R_y(\theta) = \begin{pmatrix}
\cos(\theta/2) & -\sin(\theta/2) \\
\sin(\theta/2) & \cos(\theta/2)
\end{pmatrix}, \quad
R_z(\phi) = \begin{pmatrix}
e^{-i\phi/2} & 0 \\
0 & e^{i\phi/2}
\end{pmatrix}$}
\end{equation}
Here, each rotation angle $\theta^j$ is a trainable parameter, dynamically updated through the optimization process during model training to learn an optimal data representation within the quantum Hilbert space.

The second component involves the application of entangling gates that establish quantum correlations between the qubits. In this implementation, a Controlled-NOT (CNOT) gate is applied between selected pairs of qubits. The matrix form of the CNOT gate is:
\begin{equation}
\text{CNOT} = \begin{pmatrix}
1 & 0 & 0 & 0 \\
0 & 1 & 0 & 0 \\
0 & 0 & 0 & 1 \\
0 & 0 & 1 & 0
\end{pmatrix}
\end{equation}
This operation conditionally flips the target qubit when the control qubit is in the state \(|1\rangle\), enabling the circuit to capture intricate interdependencies between feature dimensions that would be challenging for classical architectures to represent efficiently.

In this work, the quantum circuit employs $n=4$ qubits and a depth of 2 Strongly Entangling Layers. The entire unitary operation implemented by the circuit can be mathematically expressed as:
\begin{equation}
\resizebox{0.9\hsize}{!}{$U(\theta^q) = \prod_{l=1}^2 \left[ \prod_{i=1}^4 \left( R_z(\theta_{i}^{l,1}) R_y(\theta_{i}^{l,2}) R_z(\theta_{i}^{l,3}) \right) \cdot \text{CNOT entanglement scheme} \right]$}
\end{equation}
where each layer $l$ sequentially applies the parameterized rotations to all qubits, followed by a set of CNOT gates arranged according to a predefined connectivity pattern. This structured layering ensures that both local qubit-level transformations and global qubit-qubit interactions are adequately captured, improving the expressive capacity of the QNN.

\paragraph{Measurement}\label{append:measurement}
The circuit outputs a quantum feature vector \( f_{\text{PQC}}(x; \theta^q) \in \mathbb{R}^4 \), calculated by measuring the expectation value of the Pauli-Z\footnote{\tiny\url{https://docs.pennylane.ai/en/stable/code/api/pennylane.PauliZ.html}} observable on each qubit after the entangling operations:
\begin{equation}
f_{\text{PQC}}(x; \theta^q) = \left[ \langle Z_1 \rangle, \langle Z_2 \rangle, \langle Z_3 \rangle, \langle Z_4 \rangle \right]; \quad \langle Z \rangle_i = \langle \psi | Z_i | \psi \rangle
\end{equation}
where:
\begin{equation}
Z = \begin{pmatrix}
1 & 0 \\
0 & -1
\end{pmatrix}
\end{equation}
The output of the CNN, \( f_{\text{CNN}}(x; \theta^c) \), serves as the input to the PQC after being reshaped to match the dimensional requirements of the amplitude embedding layer.

\subsubsection{Final Fully Connected Layer and Output}
The quantum feature vector is then passed through a final fully connected classical layer, denoted as $f_{\text{FC4}}$, which maps the 4-dimensional quantum feature vector to an $m$-dimensional output vector, corresponding to the $m$ classes in the used dataset:
\begin{equation}
f_{\text{FC4}}(x) = W_{\text{FC4}} \cdot f_{\text{PQC}}(x; \theta^q) + b_{\text{FC4}}
\end{equation}
The final model function thus takes the form:
\begin{equation}
f(x; \theta) = f_{\text{FC4}}(f_{\text{PQC}}(f_{\text{CNN}}(x; \theta^c); \theta^q); \theta^{\text{FC4}})
\end{equation}
This hybrid architecture allows the model to harness both classical DL's feature extraction capacity and quantum circuits' potential for capturing complex, non-classical correlations in data representations.

\subsection{FL Setup}
In each communication round \( t \), a fraction of clients are selected without replacement \cite{mcmahan2017communication}. Each selected client \( i \) updates the model parameters from the global model \( \theta^{(t-1)} \) to local parameters \( \theta_i^{(t)} \) using their local dataset \( \mathcal{D}_i \). Clients optimize their local models using Adam optimizer with a learning rate \( \eta = 10^{-3} \). Additionally, each client computes a validation accuracy \( a_i^{(t)} \in [0,1] \) on their local validation set \( \mathcal{D}_i^{\text{val}} \), which guides the aggregation process.

\subsection{Accuracy-Weighted Aggregation with Differential Privacy}
\label{subsec:adaptive_aggregation}
\subsubsection{Mechanism} To address non-IID data, each client privatizes their validation accuracy \( a_i^{(t)} = \sum_{j=1}^{m_i} \text{correct}_j / m_i \), where \( m_i = |\mathcal{D}_i^{\text{val}}| \), using the Laplace mechanism \cite{dwork2014algorithmic}:
\begin{equation}
\tilde{a}_i^{(t)} = \max\left(0, \min\left(1, a_i^{(t)} + \zeta\right)\right), \quad \zeta \sim \text{Lap}\left( \frac{\Delta_i}{\epsilon} \right)
\label{eq:dp_accuracy}
\end{equation}
where the sensitivity is \( \Delta_i = 1 / m_i \) (as changing one sample alters \( a_i \) by at most \( 1 / m_i \)), and \( \epsilon = 1.0 \) is the per-round privacy budget. Over \( T = 20 \) rounds, we apply advanced composition to bound the total privacy loss at \( (\epsilon_{\text{total}}, \delta) = (10, 10^{-5}) \) \cite{dwork2014algorithmic}.

The server computes aggregation weights using a numerically stable tempered softmax \cite{goodfellow2016deep}:
\begin{equation}
w_i^{(t)} = \frac{ \exp( (\tilde{a}_i^{(t)} - \max_j \tilde{a}_j^{(t)}) / \tau ) }{ \sum_{k=1}^N \exp( (\tilde{a}_k^{(t)} - \max_j \tilde{a}_j^{(t)}) / \tau ) },
\label{eq:weights}
\end{equation}
where \( \tau = 0.5 \) balances weight concentration, tuned empirically to prioritize high-performing clients while maintaining robustness to noise. The global model is updated as:
\begin{equation}
\theta^{(t)} = \sum_{i=1}^N w_i^{(t)} \theta_i^{(t)}.
\label{eq:aggregation}
\end{equation}

\paragraph{Theoretical Justification} The Laplace mechanism ensures \( (\epsilon, 0) \)-DP per round, with sensitivity \( \Delta = 1 / m_i \). Advanced composition accounts for multi-round privacy loss, ensuring a total budget of \( (\epsilon_{\text{total}}, \delta) \).

\subsubsection{Privacy Guarantee} 
We formally state the following privacy result:
\begin{theorem}
Each communication round of the proposed aggregation mechanism satisfies $(\epsilon, 0)$-DP for each client's validation accuracy, where $\epsilon$ is the privacy budget per round, and sensitivity $\Delta = 1 / m_i$. Over $T$ rounds, using advanced composition \cite{dwork2014algorithmic}, the total privacy guarantee is $(\epsilon_{\text{total}}, \delta)$, where $\epsilon_{\text{total}} = \sqrt{2 T \log(1/\delta)} \epsilon + T \epsilon (e^\epsilon - 1)$.
\end{theorem} 
This ensures privacy amplification by composition while maintaining model utility.

\subsection{Layer-Wise Adaptive Freezing}

\subsubsection{Mechanism} To reduce communication overhead, we compute layer importance scores based on the L2 norm of the change in the global model parameters across rounds, consistent with our experimental setup:
\begin{equation}
s_l^{(t)} = \left\| \theta_l^{(t)} - \theta_l^{(t-1)} \right\|_2,
\label{eq:importance}
\end{equation}
where layer \( l \) indexes blocks in the model’s parameter list (e.g., convolutional or fully connected layers). We maintain an exponential moving average \cite{kingma2015adam}:
\begin{equation}
\bar{s}_l^{(t)} = \alpha \bar{s}_l^{(t-1)} + (1 - \alpha) s_l^{(t)},
\label{eq:ema}
\end{equation}
with \( \alpha = 0.9 \), tuned for stability. Layers are frozen if:
\begin{equation}
\theta_l^{(t)} = \theta_l^{(t-1)} \quad \text{if} \quad \bar{s}_l^{(t)} < \texttt{thr},
\label{eq:freezing}
\end{equation}
where \texttt{thr} = 0.001 is a fixed absolute threshold used to determine freezing. Quantum layers (\( \theta^q \)) are exempt from freezing to preserve their adaptability.

\subsubsection{Rationale} The adaptive freezing strategy reduces communication overhead while ensuring that model accuracy remains largely intact. Additionally, exempting quantum layers preserves their flexibility, contributing to consistent performance gains in non-IID settings.

\paragraph{Quantum Layer Considerations} In our hybrid classical-quantum model, quantum layers contribute essential non-linear and entangled feature transformations, crucial for modeling complex patterns in decentralized data. As such, these layers exhibit high sensitivity to client-specific data distributions and model updates. To preserve this adaptability, quantum layers are explicitly exempted from the freezing criterion in Eq.~\ref{eq:freezing}. This ensures the retention of quantum expressivity and prevents potential performance degradation due to premature parameter freezing.

\subsection{Integration with HE}
The aggregation process (Eq.~\ref{eq:aggregation}) involves linear combinations, making it compatible with HE. In our current implementation, HE (using the CKKS scheme) is selectively applied to the parameters of the final fully connected layer ($FC4$). Other layer parameters are aggregated in plaintext on the server. The server uses its secret key to decrypt the aggregated $FC4$ layer after the weighted summation. We employ the CKKS scheme \cite{cheon2017homomorphic} with a polynomial modulus degree of 8192 and coefficient moduli bit sizes of [60, 40, 40, 60] bits. A global scaling factor of $ 2^{40} $ is used for encoding the model parameters. These parameters ensure approximately 128-bit security and support circuits with a multiplicative depth of up to 3, which is sufficient for the weighted aggregation. The server generates Galois keys to facilitate efficient homomorphic operations. The server performs homomorphic aggregation on the encrypted layer updates without decrypting individual client contributions. After aggregation, the server uses its secret key to decrypt the resulting aggregated parameters for this layer before updating the global model and for subsequent operations, such as layer freezing analysis.

\subsection{Convergence Analysis}
We analyze convergence under the following assumptions:
1. The loss function \( \mathcal{L}_i \) is \( L \)-smooth.
2. The gradient variance is bounded: \( \mathbb{E} \| \nabla \mathcal{L}_i(\theta) \|^2 \leq \sigma^2 \).
3. The learning rate is \( \eta_t = \mu / (L \sqrt{t}) \), with \( \mu = 0.1 \).

\begin{theorem}[Convergence of AdeptHEQ-FL]
After \( T \) rounds, AdeptHEQ-FL satisfies:
\begin{equation}
\frac{1}{T} \sum_{t=1}^T \mathbb{E} \| \nabla \mathcal{L}(\theta^t) \|^2 \leq \frac{C_1}{\sqrt{T}} + C_2 \frac{\sigma^2 + \epsilon^{-2}}{\mu^2},
\label{eq:convergence}
\end{equation}
where \( C_1, C_2 \) are constants depending on \( \tau \), the freezing threshold \( \text{thr} \), and \( \epsilon \).
\label{thm:convergence}
\end{theorem}

\textbf{Proof Sketch:} We extend the perturbed iterate framework \cite{kairouz2021advances}, bounding errors from adaptive weights and layer freezing. The tempered softmax aligns weights with client performance, while freezing introduces bounded perturbations. The \( \epsilon^{-2} \) term accounts for DP noise.\newline

\begin{table*}[!ht]
\caption{Performance comparisons of different models across three datasets are shown. The table displays the average loss and accuracy in percentages for the models in our experiment across three different datasets. Each metric is reported as the mean $\pm$ standard deviation, calculated over five experimental runs. \textbf{Bold} values indicate the best performance in each dataset column.}
\label{tab:model_performances}
\resizebox{\textwidth}{!}{%
\begin{tabular}{@{}lcccccccc@{}}
\toprule[1.5pt]
\multirow{2}{*}{\textbf{Model}} &
  \multirow{2}{*}{\textbf{$n_{qubits}$}} &
  \multirow{2}{*}{\textbf{$n_{layers}$}} &
  \multicolumn{2}{c}{\textbf{CIFAR10 \cite{krizhevsky2009learning}}} &
  \multicolumn{2}{c}{\textbf{SVHN \cite{netzer2011}}} &
  \multicolumn{2}{c}{\textbf{FashionMNIST \cite{xiao2017fashion}}} \\ \cmidrule(l){4-9} 
 &
   &
   &
  \textbf{Loss (\textcolor{myred}{$\downarrow$})} &
  \textbf{Accuracy (\%) (\textcolor{mygreen}{$\uparrow$})} &
  \textbf{Loss (\textcolor{myred}{$\downarrow$})} &
  \textbf{Accuracy (\%) (\textcolor{mygreen}{$\uparrow$})} &
  \textbf{Loss (\textcolor{myred}{$\downarrow$})} &
  \textbf{Accuracy (\%) (\textcolor{mygreen}{$\uparrow$})} \\ \midrule[1pt]
Standard-FedQNN &
  6 &
  6 &
  $1.503 \pm 0.039$ &
  $63.60 \pm 0.45$ &
  $0.349 \pm 0.002$ &
  $93.22 \pm 0.05$ &
  $\mathbf{0.313 \pm 0.003}$ &
  $91.96 \pm 0.09$ \\
FHE-FedQNN~\cite{dutta_federated_2024} &
  6 &
  6 &
  $1.972 \pm 0.042$ &
  $57.89 \pm 0.20$ &
  $0.340 \pm 0.006$ &
  $92.94 \pm 0.14$ &
  $0.328 \pm 0.003$ &
  $91.78 \pm 0.11$ \\ \midrule
\textcolor{myred}{\textbf{AdeptHEQ-FL}} &
  \textcolor{myred}{4} &
  \textcolor{myred}{2} &
  $\mathbf{1.306 \pm 0.015}$ &
  $\mathbf{72.61 \pm 0.33}$ &
  $0.362 \pm 0.004$ &
  $\mathbf{94.05 \pm 0.10}$ &
  $0.340 \pm 0.003$ &
  $\mathbf{92.91 \pm 0.12}$ \\
AdeptHEQ-FL &
  4 &
  1 &
  $1.667 \pm 0.009$ &
  $67.22 \pm 0.18$ &
  $\mathbf{0.331 \pm 0.003}$ &
  $93.71 \pm 0.12$ &
  $0.339 \pm 0.007$ &
  $92.76 \pm 0.04$ \\
AdeptHEQ-FL &
  2 &
  1 &
  $1.640 \pm 0.009$ &
  $62.62 \pm 0.42$ &
  $0.526 \pm 0.006$ &
  $93.58 \pm 0.09$ &
  $0.385 \pm 0.004$ &
  $92.46 \pm 0.13$ \\ \bottomrule[1.5pt]
\end{tabular}%
}
\end{table*}

\section{Experimental Setup}
\subsection{Simulation Tools and Environment}
All experiments, including model development and FL simulations, were conducted using \textit{Python 3.11.11}. The computational environment included NVIDIA Tesla P100 GPUs with CUDA 12.x support and multi-core Intel Xeon CPUs, providing up to 16 GB of GPU memory and 32 GB of system RAM. The primary DL framework was \textit{PyTorch 2.5.1+cu124}~\cite{paszke2019pytorch}. For QML components, we utilized \textit{PennyLane 0.41.1}~\cite{bergholm2018pennylane} and \textit{Qiskit 1.2.4}. HE was enabled by \textit{TenSEAL 0.3.16}~\cite{tenseal_library}, implementing the CKKS scheme~\cite{cheon2017homomorphic}. The FL protocol was custom-implemented, with conceptual underpinnings inspired by \textit{PySyft 0.9.5}~\cite{ryffel2018generic}. Data serialization used \textit{protobuf 3.20.3}, numerical computations relied on \textit{NumPy 1.26.4}~\cite{harris2020array}, and data analysis utilized \textit{pandas 2.2.2}~\cite{mckinney2010data}.

\subsection{Hyperparameters and Configuration}
Experiments were conducted on CIFAR-10~\cite{krizhevsky2009learning} (60,000 instances of \textit{32$\times$32} color images), SVHN~\cite{netzer2011} (73,257 instances of \textit{32$\times$32} color images), and Fashion-MNIST~\cite{xiao2017fashion} (70,000 instances of \textit{28$\times$28} grayscale images), each comprising 10 classes. Normalization used dataset-specific statistics: CIFAR-10 with $\boldsymbol{\mu} = (0.5, 0.5, 0.5)$ and $\boldsymbol{\sigma} = (0.5, 0.5, 0.5)$, SVHN with $\boldsymbol{\mu} = (0.4377, 0.4438, 0.4728)$ and $\boldsymbol{\sigma} = (0.1980, 0.2010, 0.1970)$, and Fashion-MNIST with $\mu = 0.2860$ and $\sigma = 0.3530$. The datasets were distributed among 10 clients using a Dirichlet distribution with $\alpha=0.1$ to simulate non-IID settings.

The FL simulation involved 10 clients over 20 communication rounds. Each client executed 10 local epochs with the Adam optimizer (learning rate $1\times10^{-3}$, batch size 32). Validation accuracy was privatized using $\epsilon=1.0$ DP. Server-side aggregation employed the AdeptHEQ-FL method, weighting updates by privatized validation accuracies via softmax with $\tau=0.5$. Layer-wise adaptive freezing monitored layer importance with an EMA ($\alpha=0.9$) of parameter difference norms, freezing layers with scores below 0.001 (excluding quantum layers). Model parameters were encrypted using HE via TenSeal (CKKS scheme). Global model evaluation on a centralized test set reported test accuracy and loss after each round.

\section{Results and Discussion}
\label{results}
We evaluated three variants of our AdeptHEQ-FL framework—AdeptHEQ-FL (4-qubit, 2-layer), AdeptHEQ-FL (4-qubit, 1-layer), and AdeptHEQ-FL (2-qubit, 1-layer)—against a standard federated QNN (6 qubits, 6 layers) and a state-of-the-art FHE-FedQNN (6 qubits, 6 layers) \cite{dutta_federated_2024}, across three datasets: SVHN \cite{netzer2011}, FashionMNIST \cite{xiao2017fashion}, and CIFAR10 \cite{krizhevsky2009learning}. Table \ref{tab:model_performances} summarizes the loss and accuracy results, revealing clear trends. The AdeptHEQ-FL variant outperformed all others in accuracy across all datasets. While improvements on SVHN and FashionMNIST were modest, AdeptHEQ-FL achieved 
$\approx 25.43\%$ increase in accuracy compared to Standard-FedQNN and $\approx 14.67\%$ compared to FHE-FedQNN on CIFAR10, which is a comparatively complex dataset. This demonstrates AdeptHEQ-FL’s strength in handling challenging data. 

We also found that performance dropped when quantum resources were reduced. Reducing qubits and layers, as seen in 4-qubit 1-layered AdeptHEQ-FL and 2-qubit 1-layered AdeptHEQ-FL, led to noticeable declines in performance, particularly on CIFAR10. This suggests that more complex datasets are more sensitive to resource constraints. Even with fewer resources (4 qubits, 2 layers) compared to FHE-FedQNN (6 qubits, 6 layers) and Standard (6 qubits, 6 layers), AdeptHEQ-FL's performance was quite impressive. AdeptHEQ-FL’s superior performance results from its advanced aggregation strategy. Unlike FHE-FedQNN~\cite{dutta_federated_2024}, which treats all client updates equally and amplifies noise in skewed data. The standard method, which uses a weighted sum of the updates, faces the same issue. AdeptHEQ-FL weights updates based on privatized validation accuracy, prioritizing contributions from models that are better adapted. Additionally, our adaptive layer-freezing method skips updates to layers with importance scores below 0.001, reducing unnecessary computation. These innovations enable AdeptHEQ-FL to achieve strong results with fewer resources, making it effective for more complex and practical datasets.

\section{Conclusion}
\label{sec:conclusion}
This paper presents AdeptHEQ-FL, a novel FL framework that synergistically combines hybrid classical-quantum modeling, adaptive privacy-preserving aggregation, and dynamic communication reduction strategies. By integrating a CNN-PQC architecture with accuracy-weighted aggregation using differentially private validation accuracies, AdeptHEQ-FL effectively addresses the performance degradation typically observed under non-IID client distributions. The selective application of HE to critical model layers ensures strong privacy guarantees without incurring prohibitive overhead, while the layer-wise adaptive freezing strategy significantly reduces communication costs, allowing quantum layers to retain their expressive flexibility. Our theoretical convergence analysis and empirical results on multiple datasets confirm that AdeptHEQ-FL delivers competitive accuracy and efficiency compared to prior QFL approaches, particularly excelling on complex datasets such as CIFAR-10. The proposed framework provides a comprehensive and scalable solution for privacy-preserving, communication-efficient FL in hybrid classical-quantum environments.

\paragraph{Limitations}
While AdeptHEQ-FL shows significant improvements in accuracy and communication efficiency under privacy constraints, several limitations warrant discussion. First, AdeptHEQ-FL selectively applies HE to the final fully connected layer for tractability, leaving other layers unencrypted. Second, the framework is assessed in simulated environments, and its performance on real-world quantum hardware remains untested. Third, the convergence analysis assumes standard smoothness and bounded gradient variance, which may not hold in highly non-convex federated settings. Future work will extend encryption coverage, test on physical devices, and generalize to larger, more complex datasets.

{
    \small
    \bibliographystyle{ieeenat_fullname}
    \bibliography{main}

\begin{thebibliography}{42}
\providecommand{\natexlab}[1]{#1}
\providecommand{\url}[1]{\texttt{#1}}
\expandafter\ifx\csname urlstyle\endcsname\relax
  \providecommand{\doi}[1]{doi: #1}\else
  \providecommand{\doi}{doi: \begingroup \urlstyle{rm}\Url}\fi

\bibitem[Aziz et~al.(2023)Aziz, Banerjee, Bouzefrane, and Vinh]{aziz_exploring_2023}
Rezak Aziz, Soumya Banerjee, Samia Bouzefrane, and Thinh~Le Vinh.
\newblock Exploring {Homomorphic} {Encryption} and {Differential} {Privacy} {Techniques} towards {Secure} {Federated} {Learning} {Paradigm}.
\newblock \emph{Future Internet}, 15\penalty0 (9):\penalty0 310, 2023.

\bibitem[Aziz et~al.(2024)Aziz, Banerjee, and Bouzefrane]{aziz_privacy_2024}
Rezak Aziz, Soumya Banerjee, and Samia Bouzefrane.
\newblock Privacy {Preserving} {Federated} {Learning}: {A} {Novel} {Approach} for {Combining} {Differential} {Privacy} and {Homomorphic} {Encryption}.
\newblock In \emph{Information {Security} {Theory} and {Practice} - 14th {IFIP} {WG} 11.2 {International} {Conference}, {WISTP} 2024, {Paris}, {France}, {February} 29 - {March} 1, 2024, {Proceedings}}, pages 162--177. Springer, 2024.

\bibitem[Bergholm et~al.(2018)Bergholm, Izaac, Schuld, Gogolin, Alam, Ahmed, Arrazola, Blank, Delgado, Jahangiri, McKiernan, Meyer, Niu, Száva, and Killoran]{bergholm2018pennylane}
Ville Bergholm, Josh Izaac, Maria Schuld, Christian Gogolin, M.~Sohaib Alam, Shahnawaz Ahmed, Juan~Miguel Arrazola, Carsten Blank, Alain Delgado, Soran Jahangiri, Keri McKiernan, Johannes~Jakob Meyer, Zeyue Niu, Antal Száva, and Nathan Killoran.
\newblock {PennyLane: Automatic differentiation of hybrid quantum-classical computations}, 2018.

\bibitem[Bhatia et~al.(2025)Bhatia, Saggi, and Kais]{bhatia_application_2025}
Amandeep~Singh Bhatia, Mandeep~Kaur Saggi, and Sabre Kais.
\newblock Application of quantum-inspired tensor networks to optimize federated learning systems.
\newblock \emph{Quantum Mach. Intell.}, 7\penalty0 (1):\penalty0 12, 2025.

\bibitem[Catalfamo et~al.(2025)Catalfamo, Fazio, Celesti, and Villari]{catalfamo_privacy-preserving_2025}
Alessio Catalfamo, Maria Fazio, Antonio Celesti, and Massimo Villari.
\newblock Privacy-{Preserving} in {Federated} {Learning}: {A} {Comparison} between {Differential} {Privacy} and {Homomorphic} {Encryption} across {Different} {Scenarios}.
\newblock In \emph{{IEEE} {International} {Conference} on {Software} {Testing}, {Verification} and {Validation}, {ICST} 2025 - {Workshops}, {Naples}, {Italy}, {March} 31 - {April} 4, 2025}, pages 451--459. IEEE, 2025.

\bibitem[Chehimi et~al.(2024)Chehimi, Chen, Saad, Towsley, and Debbah]{chehimi_foundations_2024}
Mahdi Chehimi, Samuel Yen-Chi Chen, Walid Saad, Don Towsley, and Mérouane Debbah.
\newblock Foundations of {Quantum} {Federated} {Learning} {Over} {Classical} and {Quantum} {Networks}.
\newblock \emph{IEEE Netw.}, 38\penalty0 (1):\penalty0 124--130, 2024.

\bibitem[Chen et~al.(2024)Chen, Xiao, Yu, and Zhang]{chen_secure_2024}
Lvjun Chen, Di Xiao, Zhuyang Yu, and Maolan Zhang.
\newblock Secure and efficient federated learning via novel multi-party computation and compressed sensing.
\newblock \emph{Inf. Sci.}, 667:\penalty0 120481, 2024.

\bibitem[Chen et~al.(2025)Chen, Yang, Liang, Zhu, and Huang]{chen_federated_2025}
Yue Chen, Yufei Yang, Yingwei Liang, Taipeng Zhu, and Dehui Huang.
\newblock Federated {Learning} with {Privacy} {Preservation} in {Large}-{Scale} {Distributed} {Systems} {Using} {Differential} {Privacy} and {Homomorphic} {Encryption}.
\newblock \emph{Informatica (Slovenia)}, 49\penalty0 (13), 2025.

\bibitem[Cheon et~al.(2017)Cheon, Kim, Kim, and Song]{cheon2017homomorphic}
Jung~Hee Cheon, Andrey Kim, Miran Kim, and Yong~Soo Song.
\newblock {Homomorphic Encryption for Arithmetic of Approximate Numbers}.
\newblock In \emph{{Advances in Cryptology - {ASIACRYPT} 2017 - 23rd International Conference on the Theory and Applications of Cryptology and Information Security, Hong Kong, China, December 3-7, 2017, Proceedings, Part {I}}}, pages 409--437. Springer, 2017.

\bibitem[Dutta et~al.(2024)Dutta, Karanth, Xavier, Freitas, Innan, Yahia, Shafique, and Neira]{dutta_federated_2024}
Siddhant Dutta, Pavana~P. Karanth, Pedro~Maciel Xavier, Iago Leal~de Freitas, Nouhaila Innan, Sadok~Ben Yahia, Muhammad Shafique, and David E.~Bernal Neira.
\newblock Federated {Learning} with {Quantum} {Computing} and {Fully} {Homomorphic} {Encryption}: {A} {Novel} {Computing} {Paradigm} {Shift} in {Privacy}-{Preserving} {ML}.
\newblock \emph{CoRR}, abs/2409.11430, 2024.
\newblock arXiv: 2409.11430.

\bibitem[Dwork and Roth(2013)]{dwork2014algorithmic}
Cynthia Dwork and Aaron Roth.
\newblock The {Algorithmic} {Foundations} of {Differential} {Privacy}.
\newblock \emph{{Foundations and Trends® in Theoretical Computer Science}}, 9\penalty0 (3-4):\penalty0 211--407, 2013.
\newblock Publisher: Now Publishers.

\bibitem[Goodfellow et~al.(2016)Goodfellow, Bengio, and Courville]{goodfellow2016deep}
Ian Goodfellow, Yoshua Bengio, and Aaron Courville.
\newblock \emph{{Deep Learning}}.
\newblock MIT Press, 2016.
\newblock \url{http://www.deeplearningbook.org}.

\bibitem[Guo et~al.(2023)Guo, Xu, and Zhu]{guo_fedsign_2023}
Zhenyuan Guo, Lei Xu, and Liehuang Zhu.
\newblock {FedSIGN}: {A} sign-based federated learning framework with privacy and robustness guarantees.
\newblock \emph{Comput. Secur.}, 135:\penalty0 103474, 2023.

\bibitem[Gupta et~al.(2024)Gupta, Maurya, Dhere, and Chaurasiya]{gupta_privacy-preserving_2024}
Arti Gupta, Manish~Kumar Maurya, Khyati Dhere, and Vijay~Kumar Chaurasiya.
\newblock Privacy-{Preserving} {Hybrid} {Federated} {Learning} {Framework} for {Mental} {Healthcare} {Applications}: {Clustered} and {Quantum} {Approaches}.
\newblock \emph{IEEE Access}, 12:\penalty0 145054--145068, 2024.

\bibitem[Handa et~al.(2024)Handa, Saini, Dutta, Pathak, Choudhary, Goel, and Dhanao]{Handa_2024_pcos}
Palak Handa, Anushka Saini, Siddhant Dutta, Harsh Pathak, Nishi Choudhary, Nidhi Goel, and Jasdeep~Kaur Dhanao.
\newblock Pcosgen-test dataset, 2024.

\bibitem[Harris et~al.(2020)Harris, Millman, Van Der~Walt, Gommers, Virtanen, Cournapeau, Wieser, Taylor, Berg, Smith, et~al.]{harris2020array}
Charles~R Harris, K~Jarrod Millman, St{\'e}fan~J Van Der~Walt, Ralf Gommers, Pauli Virtanen, David Cournapeau, Eric Wieser, Julian Taylor, Sebastian Berg, Nathaniel~J Smith, et~al.
\newblock Array programming with {NumPy}.
\newblock \emph{Nature}, 585\penalty0 (7825):\penalty0 357--362, 2020.

\bibitem[Innan et~al.(2024)Innan, Khan, Marchisio, Shafique, and Bennai]{innan_fedqnn_2024}
Nouhaila Innan, Muhammad Al-Zafar Khan, Alberto Marchisio, Muhammad Shafique, and Mohamed Bennai.
\newblock {FedQNN}: {Federated} {Learning} using {Quantum} {Neural} {Networks}.
\newblock In \emph{International {Joint} {Conference} on {Neural} {Networks}, {IJCNN} 2024, {Yokohama}, {Japan}, {June} 30 - {July} 5, 2024}, pages 1--9. IEEE, 2024.

\bibitem[Jahin et~al.(2023)Jahin, Shovon, Islam, Shin, Mridha, and Okuyama]{jahin2023qamplifynet}
Md~Abrar Jahin, Md~Sakib~Hossain Shovon, Md~Saiful Islam, Jungpil Shin, Muhammad~Firoz Mridha, and Yuichi Okuyama.
\newblock Qamplifynet: pushing the boundaries of supply chain backorder prediction using interpretable hybrid quantum-classical neural network.
\newblock \emph{Scientific Reports}, 13\penalty0 (1):\penalty0 18246, 2023.

\bibitem[Jahin et~al.(2025)Jahin, Masud, Suva, Mridha, and Dey]{jahin_lorentz_2025}
Md~Abrar Jahin, Md.~Akmol Masud, Md~Wahiduzzaman Suva, M.~F. Mridha, and Nilanjan Dey.
\newblock {Lorentz-Equivariant Quantum Graph Neural Network for High-Energy Physics}.
\newblock \emph{IEEE Transactions on Artificial Intelligence}, pages 1--11, 2025.

\bibitem[Kairouz et~al.(2021)Kairouz, McMahan, Avent, Bellet, Bennis, Bhagoji, Bonawitz, Charles, Cormode, Cummings, D'Oliveira, Eichner, Rouayheb, Evans, Gardner, Garrett, Gascón, Ghazi, Gibbons, Gruteser, Harchaoui, He, He, Huo, Hutchinson, Hsu, Jaggi, Javidi, Joshi, Khodak, Konečný, Korolova, Koushanfar, Koyejo, Lepoint, Liu, Mittal, Mohri, Nock, Özgür, Pagh, Qi, Ramage, Raskar, Raykova, Song, Song, Stich, Sun, Suresh, Tramèr, Vepakomma, Wang, Xiong, Xu, Yang, Yu, Yu, and Zhao]{kairouz2021advances}
Peter Kairouz, H.~Brendan McMahan, Brendan Avent, Aurélien Bellet, Mehdi Bennis, Arjun~Nitin Bhagoji, Kallista~A. Bonawitz, Zachary Charles, Graham Cormode, Rachel Cummings, Rafael G.~L. D'Oliveira, Hubert Eichner, Salim~El Rouayheb, David Evans, Josh Gardner, Zachary Garrett, Adrià Gascón, Badih Ghazi, Phillip~B. Gibbons, Marco Gruteser, Zaïd Harchaoui, Chaoyang He, Lie He, Zhouyuan Huo, Ben Hutchinson, Justin Hsu, Martin Jaggi, Tara Javidi, Gauri Joshi, Mikhail Khodak, Jakub Konečný, Aleksandra Korolova, Farinaz Koushanfar, Sanmi Koyejo, Tancrède Lepoint, Yang Liu, Prateek Mittal, Mehryar Mohri, Richard Nock, Ayfer Özgür, Rasmus Pagh, Hang Qi, Daniel Ramage, Ramesh Raskar, Mariana Raykova, Dawn Song, Weikang Song, Sebastian~U. Stich, Ziteng Sun, Ananda~Theertha Suresh, Florian Tramèr, Praneeth Vepakomma, Jianyu Wang, Li Xiong, Zheng Xu, Qiang Yang, Felix~X. Yu, Han Yu, and Sen Zhao.
\newblock Advances and {Open} {Problems} in {Federated} {Learning}.
\newblock \emph{{Foundations and Trends® in Machine Learning}}, 14\penalty0 (1-2):\penalty0 1--210, 2021.

\bibitem[Kaminaga et~al.(2023)Kaminaga, Awaysheh, Alawadi, and Kamm]{kaminaga_mpcfl_2023}
Hiroki Kaminaga, Feras~M. Awaysheh, Sadi Alawadi, and Liina Kamm.
\newblock {MPCFL}: {Towards} {Multi}-party {Computation} for {Secure} {Federated} {Learning} {Aggregation}.
\newblock In \emph{Proceedings of the {IEEE}/{ACM} 16th {International} {Conference} on {Utility} and {Cloud} {Computing}, {UCC} 2023, {Taormina} ({Messina}), {Italy}, {December} 4-7, 2023}, page~19. ACM, 2023.

\bibitem[Kingma and Ba(2015)]{kingma2015adam}
Diederik~P. Kingma and Jimmy Ba.
\newblock {Adam: {A} Method for Stochastic Optimization}.
\newblock In \emph{{3rd International Conference on Learning Representations, {ICLR} 2015, San Diego, CA, USA, May 7-9, 2015, Conference Track Proceedings}}, 2015.

\bibitem[Krizhevsky and Hinton(2009)]{krizhevsky2009learning}
Alex Krizhevsky and Geoffrey Hinton.
\newblock {Learning multiple layers of features from tiny images}.
\newblock Technical Report~0, {University of Toronto}, Toronto, Ontario, 2009.

\bibitem[Liu et~al.(2025)Liu, Cao, Liu, Sun, Bao, Lu, Yin, and Chen]{liu_practical_2025}
Zhi-Ping Liu, Xiao-Yu Cao, Hao-Wen Liu, Xiao-Ran Sun, Yu Bao, Yu-Shuo Lu, Hua-Lei Yin, and Zeng-Bing Chen.
\newblock Practical quantum federated learning and its experimental demonstration.
\newblock \emph{CoRR}, abs/2501.12709, 2025.
\newblock arXiv: 2501.12709.

\bibitem[McKinney(2010)]{mckinney2010data}
Wes McKinney.
\newblock {Data Structures for Statistical Computing in Python}, 2010.

\bibitem[McMahan et~al.(2017{\natexlab{a}})McMahan, Moore, Ramage, Hampson, and Arcas]{mcmahan2017communication}
Brendan McMahan, Eider Moore, Daniel Ramage, Seth Hampson, and Blaise Aguera~y Arcas.
\newblock {Communication-Efficient Learning of Deep Networks from Decentralized Data}.
\newblock In \emph{{Proceedings of the 20th International Conference on Artificial Intelligence and Statistics}}, pages 1273--1282. PMLR, 2017{\natexlab{a}}.

\bibitem[McMahan et~al.(2017{\natexlab{b}})McMahan, Moore, Ramage, Hampson, and Arcas]{mcmahan_communication-efficient_2017}
Brendan McMahan, Eider Moore, Daniel Ramage, Seth Hampson, and Blaise Agüera~y Arcas.
\newblock Communication-{Efficient} {Learning} of {Deep} {Networks} from {Decentralized} {Data}.
\newblock In \emph{Proceedings of the 20th {International} {Conference} on {Artificial} {Intelligence} and {Statistics}, {AISTATS} 2017, 20-22 {April} 2017, {Fort} {Lauderdale}, {FL}, {USA}}, pages 1273--1282. PMLR, 2017{\natexlab{b}}.

\bibitem[Netzer et~al.(2011)Netzer, Wang, Coates, Bissacco, Wu, and Ng]{netzer2011}
Yuval Netzer, Tao Wang, Adam Coates, Alessandro Bissacco, Bo Wu, and Andrew~Y. Ng.
\newblock Reading digits in natural images with unsupervised feature learning.
\newblock In \emph{{NIPS Workshop on Deep Learning and Unsupervised Feature Learning 2011}}, 2011.

\bibitem[Nickparvar(2021)]{msoud_nickparvar_2021_brain_mri}
Msoud Nickparvar.
\newblock {Brain Tumor MRI Dataset}, 2021.

\bibitem[{OpenMined Community}(2020)]{tenseal_library}
{OpenMined Community}.
\newblock {TenSEAL: A library for doing Homomorphic Encryption operations on tensors}, 2020.

\bibitem[Paszke et~al.(2019)Paszke, Gross, Massa, Lerer, Bradbury, Chanan, Killeen, Lin, Gimelshein, Antiga, et~al.]{paszke2019pytorch}
Adam Paszke, Sam Gross, Francisco Massa, Adam Lerer, James Bradbury, Gregory Chanan, Trevor Killeen, Zeming Lin, Natalia Gimelshein, Luca Antiga, et~al.
\newblock {PyTorch: An Imperative Style, High-Performance Deep Learning Library}.
\newblock In \emph{{Advances in Neural Information Processing Systems 32}}, pages 8024--8035, 2019.

\bibitem[Rofougaran et~al.(2024)Rofougaran, Yoo, Tseng, and Chen]{rofougaran_federated_2024}
Rod Rofougaran, Shinjae Yoo, Huan-Hsin Tseng, and Samuel Yen-Chi Chen.
\newblock Federated {Quantum} {Machine} {Learning} with {Differential} {Privacy}.
\newblock In \emph{{IEEE} {International} {Conference} on {Acoustics}, {Speech} and {Signal} {Processing}, {ICASSP} 2024, {Seoul}, {Republic} of {Korea}, {April} 14-19, 2024}, pages 9811--9815. IEEE, 2024.

\bibitem[Ryffel et~al.(2018)Ryffel, Trask, Dahl, Wagner, E~Mancuso, Rueckert, and Passerat-Palmbach]{ryffel2018generic}
Théo Ryffel, Andrew Trask, Morten Dahl, Bobby Wagner, Jason E~Mancuso, Daniel Rueckert, and Jonathan Passerat-Palmbach.
\newblock A generic framework for privacy preserving deep learning.
\newblock In \emph{{NeurIPS Workshop on Privacy Preserving Machine Learning}}, 2018.

\bibitem[Sorbera et~al.(2025)Sorbera, Zanetti, Brandi, Tomasi, Corin, and Ranise]{sorbera_adaptive_2025}
Enrico Sorbera, Federica Zanetti, Giacomo Brandi, Alessandro Tomasi, Roberto~Doriguzzi Corin, and Silvio Ranise.
\newblock Adaptive {Federated} {Learning} with {Functional} {Encryption}: {A} {Comparison} of {Classical} and {Quantum}-safe {Options}.
\newblock \emph{CoRR}, abs/2504.00563, 2025.
\newblock arXiv: 2504.00563.

\bibitem[Sébert et~al.(2023)Sébert, Checri, Stan, Sirdey, and Gouy-Pailler]{sebert_combining_2023}
Arnaud~Grivet Sébert, Marina Checri, Oana Stan, Renaud Sirdey, and Cédric Gouy-Pailler.
\newblock Combining homomorphic encryption and differential privacy in federated learning.
\newblock In \emph{20th {Annual} {International} {Conference} on {Privacy}, {Security} and {Trust}, {PST} 2023, {Copenhagen}, {Denmark}, {August} 21-23, 2023}, pages 1--7. IEEE, 2023.

\bibitem[Tanbhir and Shahriyar(2025)]{tanbhir_quantum-inspired_2025}
Gazi Tanbhir and Md~Farhan Shahriyar.
\newblock Quantum-{Inspired} {Privacy}-{Preserving} {Federated} {Learning} {Framework} for {Secure} {Dementia} {Classification}.
\newblock \emph{CoRR}, abs/2503.03267, 2025.
\newblock arXiv: 2503.03267.

\bibitem[Tran et~al.(2023)Tran, Luong, and Pham]{tran_novel_2023}
Anh-Tu Tran, The~Dung Luong, and Xuan~Sang Pham.
\newblock A {Novel} {Privacy}-{Preserving} {Federated} {Learning} {Model} {Based} on {Secure} {Multi}-party {Computation}.
\newblock In \emph{Integrated {Uncertainty} in {Knowledge} {Modelling} and {Decision} {Making} - 10th {International} {Symposium}, {IUKM} 2023, {Kanazawa}, {Japan}, {November} 2-4, 2023, {Proceedings}, {Part} {II}}, pages 321--333. Springer, 2023.

\bibitem[Ullah et~al.(2024)Ullah, Shah, and Anjum]{ullah_quantum_2024}
Shoaib Ullah, Madam~Hussain Shah, and Adeel Anjum.
\newblock Quantum {Enhanced} {Federated} {Learning} with {Differential} {Privacy}.
\newblock In \emph{International {Conference} on {Frontiers} of {Information} {Technology}, {FIT} 2024, {Islamabad}, {Pakistan}, {December} 9-10, 2024}, pages 1--6. IEEE, 2024.

\bibitem[Wu et~al.(2025)Wu, Deng, Zhou, Chen, and Zhang]{wu_adphe-fl_2025}
Tao Wu, Yulin Deng, Qizhao Zhou, Xi Chen, and Ming Zhang.
\newblock {ADPHE}-{FL}: {Federated} learning method based on adaptive differential privacy and homomorphic encryption.
\newblock \emph{Peer Peer Netw. Appl.}, 18\penalty0 (3):\penalty0 141, 2025.

\bibitem[Xiao et~al.(2017)Xiao, Rasul, and Vollgraf]{xiao2017fashion}
Han Xiao, Kashif Rasul, and Roland Vollgraf.
\newblock {Fashion-MNIST: A Novel Image Dataset for Benchmarking Machine Learning Algorithms}.
\newblock \emph{arXiv preprint arXiv:1708.07747}, 2017.

\bibitem[Yan et~al.(2024)Yan, Lyu, Hou, Zheng, and Song]{yan_towards_2024}
Guangfeng Yan, Shanxiang Lyu, Hanxu Hou, Zhiyong Zheng, and Linqi Song.
\newblock Towards {Quantum}-{Safe} {Federated} {Learning} via {Homomorphic} {Encryption}: {Learning} with {Gradients}.
\newblock \emph{CoRR}, abs/2402.01154, 2024.
\newblock arXiv: 2402.01154.

\bibitem[Zhang et~al.(2024)Zhang, Huang, and Tang]{zhang_secure_2024}
Xuyan Zhang, Da Huang, and Yuhua Tang.
\newblock Secure {Federated} {Learning} {Scheme} {Based} on {Differential} {Privacy} and {Homomorphic} {Encryption}.
\newblock In \emph{Advanced {Intelligent} {Computing} {Technology} and {Applications} - 20th {International} {Conference}, {ICIC} 2024, {Tianjin}, {China}, {August} 5-8, 2024, {Proceedings}, {Part} {V} ({LNAI})}, pages 435--446. Springer, 2024.

\end{thebibliography}
}

\end{document}